%% file: main.tex
\documentclass[10pt,twocolumn,letterpaper]{article}

\usepackage{cvpr}              
\input{preamble}

\definecolor{cvprblue}{rgb}{0.21,0.49,0.74}
\usepackage[pagebackref,breaklinks,colorlinks,citecolor=cvprblue]{hyperref}

\newcommand{\blfootnote}[1]{
  \begingroup
  \renewcommand\thefootnote{}
  \footnotetext{#1}
  \endgroup
}

\title{Decoupling Fine Detail and Global Geometry for Compressed \\ Depth Map Super-Resolution}

\author{
Huan Zheng$^{*}$,
Wencheng Han$^{*}$,
Jianbing Shen$^{\dagger}$\\
 SKL-IOTSC, CIS, University of Macau
\vspace{-6mm}
}

\begin{document}
\maketitle

\begin{abstract}
Recovering high-quality depth maps from compressed sources has gained significant attention due to the limitations of consumer-grade depth cameras and the bandwidth restrictions during data transmission.
However, current methods still suffer from two challenges.
First, bit-depth compression produces a uniform depth representation in regions with subtle variations, hindering the recovery of detailed information.
Second, densely distributed random noise reduces the accuracy of estimating the global geometric structure of the scene.
To address these challenges, we propose a novel framework, termed geometry-decoupled network (GDNet), for compressed depth map super-resolution that decouples the high-quality depth map reconstruction process by handling global and detailed geometric features separately.
To be specific, we propose the fine geometry detail encoder (FGDE), which is designed to aggregate fine geometry details in high-resolution low-level image features while simultaneously enriching them with complementary information from low-resolution context-level image features.
In addition, we develop the global geometry encoder (GGE) that aims at suppressing noise and extracting global geometric information effectively via constructing compact feature representation in a low-rank space.
We conduct experiments on multiple benchmark datasets, demonstrating that our GDNet significantly outperforms current methods in terms of geometric consistency and detail recovery. 
In the ECCV 2024 AIM Compressed Depth Upsampling Challenge, our solution won the 1st place award.
Our codes are available at: \href{https://github.com/Ian0926/GDNet}{https://github.com/Ian0926/GDNet}.
\end{abstract}
\vspace{-5mm}
\blfootnote{$*$Equal contribution. $\dagger$Corresponding author: \textit{Jianbing Shen}.
This work was supported in part by the Science and Technology Development Fund of Macau SAR (FDCT) under grants 0102/2023/RIA2 and 0154/2022/A3 and 001/2024/SKL, the Jiangyin Hi-tech Industrial Development Zone under the Taihu Innovation Scheme (EF2025-00003-SKL-IOTSC), the University of Macau SRG2022-00023-IOTSC grant.}

\begin{figure}[t]
    \centering
    \subfloat[Depth map degradations during data acquisition and transmission.]{%
        \includegraphics[width=0.45\textwidth]{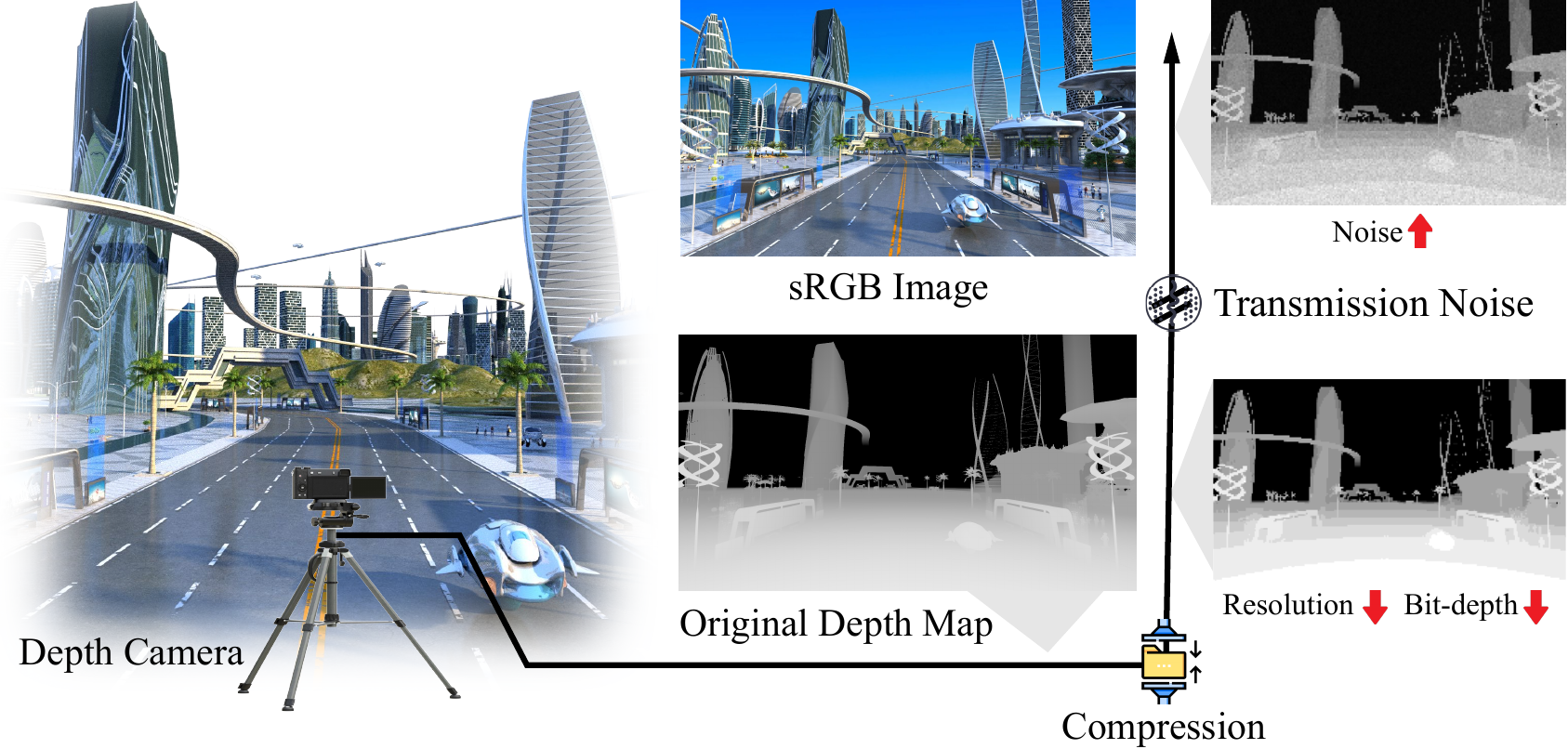}
        \vspace{-1mm}
    }  \\
    \subfloat[The motivation of the proposed GDNet.]{%
        \includegraphics[width=0.45\textwidth]{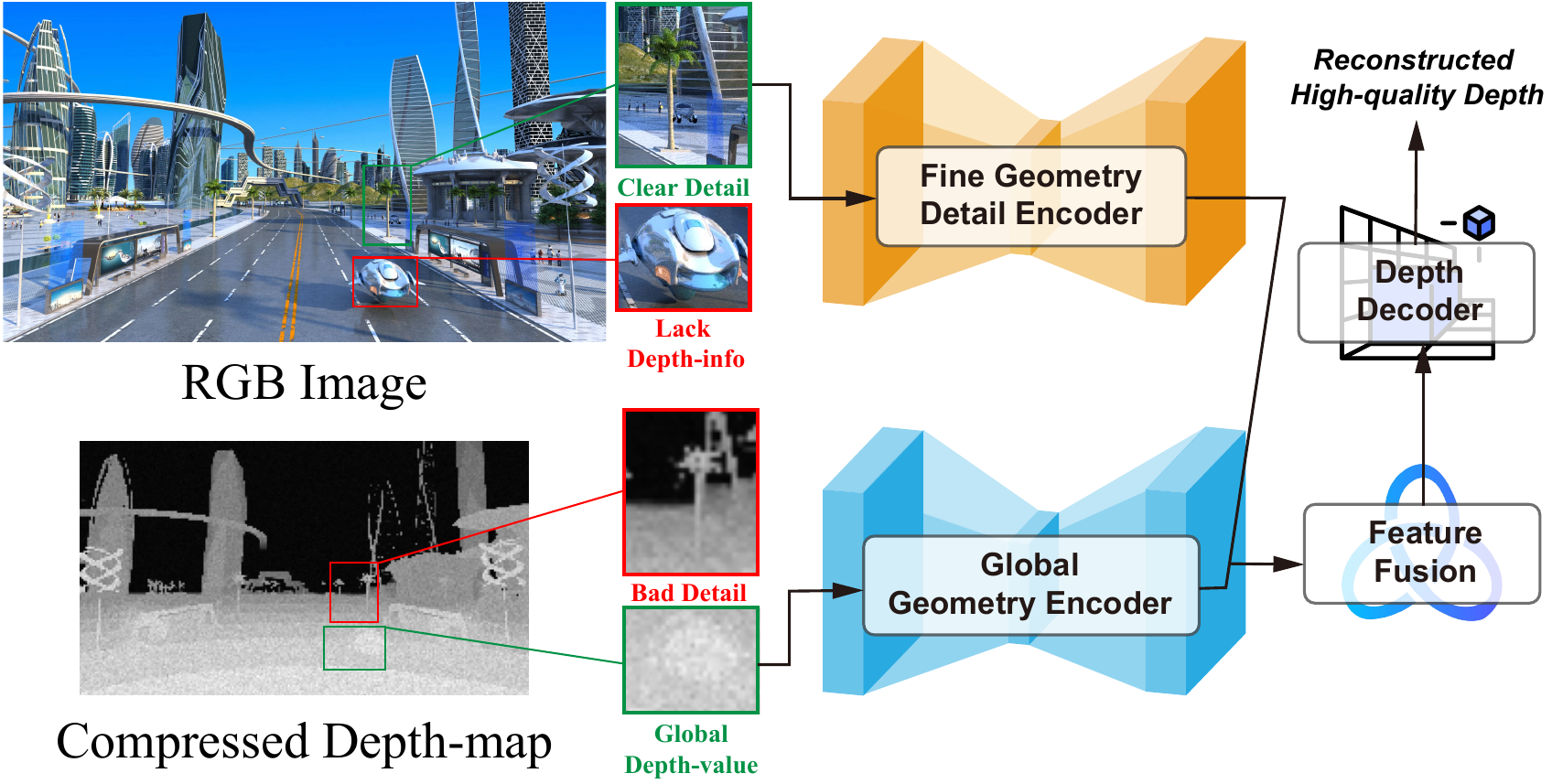}
        \vspace{-1mm}
    }
    \vspace{-2mm}
    \caption{\textbf{(a) Illustration of the Degradations during Data Acquisition and Transmission.} To be specific, the quality of the depth map is compromised due to downsampling, bit-depth compression, and noise introduced during data acquisition and transmission. \textbf{(b) The Motivation behind the Proposed GDNet.} The core idea is to leverage the compressed depth map for capturing global geometric information, while utilizing the RGB image to extract detailed geometric features. As a result, our GDNet can effectively reconstruct a high-quality depth map.
     }
    \label{fig:motivation}
    \vspace{-7mm}
\end{figure}

\section{Introduction}

Depth perception technologies play an increasingly critical role in advanced applications such as autonomous driving \cite{li2023bevdepth}, augmented reality \cite{lepetit2000semi}, and robotics \cite{dong2022towards}, where systems depend on accurate depth maps to deliver detailed 3D information for precise navigation \cite{li2022bevformer}, object recognition \cite{he2016deep}, and interaction \cite{lin2024v2vformer}. 
High-quality depth data is essential for ensuring safety \cite{wang2019pseudo}, enhancing reliability \cite{heo2018monocular}, and improving overall system performance \cite{you2019pseudo}.
However, in practical scenarios, both low-cost consumer-grade depth cameras and bandwidth-limited transmission can significantly degrade the quality of depth maps \cite{conde2024compressed}. 
Consumer-grade cameras often produce depth maps with reduced bit-depths to strike a balance between cost and performance, resulting in less accurate data \cite{wang2019comparative}. 
Moreover, under bandwidth constraints, maintaining system efficiency requires compression through downsampling and bit-depth reduction, which further degrades the quality of depth maps \cite{conde2024compressed}.
The introduction of random noise during both acquisition and transmission exacerbates these issues, making it even more difficult to recover high-quality depth maps from such compressed sources \cite{sterzentsenko2019self}.

Figure \ref{fig:motivation} (a) shows the detailed degradation process during depth data acquisition and transmission.
Two major issues are revealed during acquisition and transmission: \textit{(1) bit-depth compression leads to uniform depth representation in areas with subtle variations, making it difficult to recover fine geometry details accurately; (2) densely distributed noise in the compressed depth map negatively impacts the reasoning of global geometric information in the scene}.
However, existing guided depth map super-resolution (GDSR) primarily address challenges associated with resolution downsampling in depth maps, neglecting the additional complexities introduced by bit-depth compression and random noise \cite{song2020channel, ye2020pmbanet, ye2020depth, zhong2023guided, marivani2020multimodal, gu2019learned}. 
To further investigate this issue, we present the paired RGB image and compressed depth map in Figure \ref{fig:motivation} (b). 
We observe that the RGB image lacks depth information, while the compressed depth map exhibits poor performance in capturing fine geometry details. Conversely, the compressed depth map effectively provides global depth cues, whereas the RGB image contains rich fine geometry details. 

In this paper, we introduce geometry-decoupled network (GDNet), a new framework designed for compressed depth map super-resolution.
The primary objective of the proposed GDNet is to decouple the high-quality depth map reconstruction process into detailed geometric feature learning and global geometric feature extraction. 
Specifically, we propose the fine geometry detail encoder (FGDE), crafted to maintain fine geometry details in high-resolution low-level image features while concurrently enhancing them with supplementary information from low-resolution context-level image features.
In addition, we develop the global geometry encoder (GGE) that constructs the compressed feature representation in a low-rank space, effectively minimizing noise and facilitating the extraction of global geometric information.
To verify the effectiveness of the proposed method, we synthesis a new dataset termed Compressed-NYU, where the samples suffer from synchronous downsampling, bit-depth compression and random noise.
Through comprehensive experiments on multiple benchmarks, we demonstrate that GDNet significantly surpasses existing methods in terms of recovery quality, noise suppression, and fine geometry detail restoration.

The contributions of this paper are outlined as follows:
\begin{itemize}
	\item We present a new framework termed geometry-decoupled network (GDNet), which develops a decoupling strategy to independently learn global and detailed geometric features for compressed depth map super-resolution.
	\item We propose the fine geometry detail encoder (FGDE) designed to preserve fine geometry details in high-resolution low-level image features while enriching them with complementary information from low-resolution context-level image features.
	\item We develop the global geometry encoder (GGE) that facilitates compact feature representation in a low-rank space, enhancing noise suppression and effectively extracting global geometric information.
	\item Our model achieves state-of-the-art performance and obtains superior visual results on benchmark datasets.
\end{itemize}

\begin{figure*}[t]
    \centering
    \includegraphics[width=0.99\textwidth]{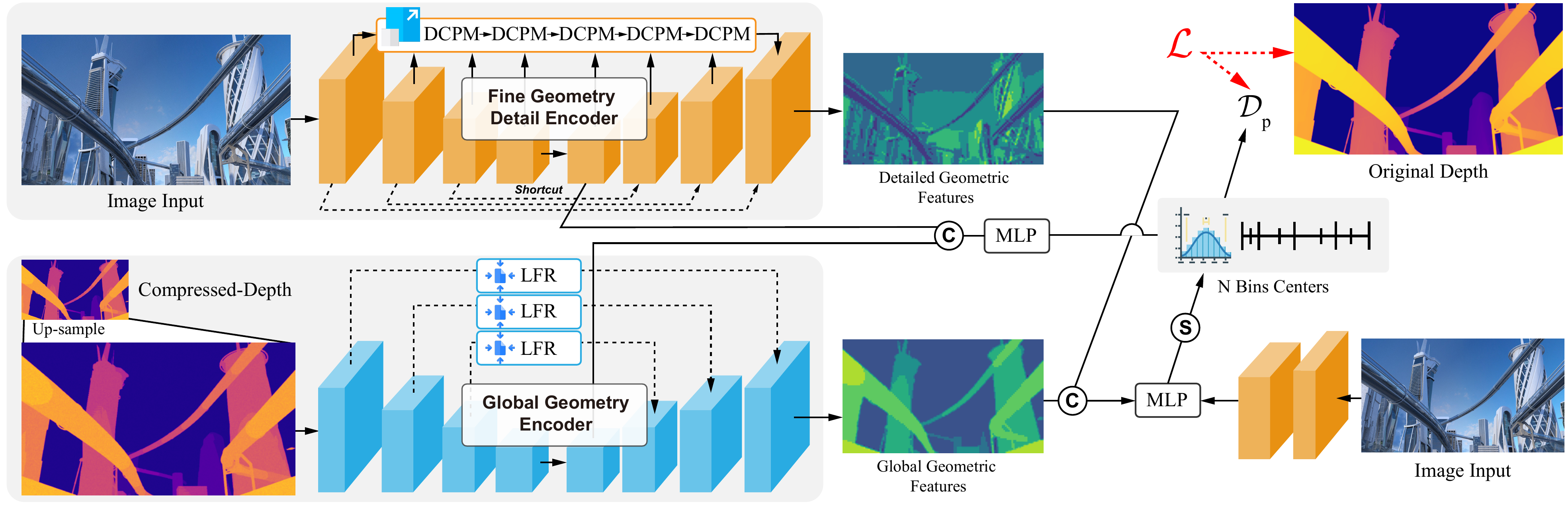}
    \vspace{-2mm}
    \caption{\textbf{The Overall Framework of the Proposed GDNet.} Our GDNet leverages RGB images to capture fine geometric details while utilizing compressed depth maps to provide global depth information. By employing the above decoupling strategy, the proposed GDNet is able to reconstruct high-quality depth maps with improved accuracy. Specifically, GDNet comprises three main components: a fine geometry detail encoder, responsible for detailed geometric feature extraction; a global geometry encoder, aiming at capturing global geometric features; a depth decoder to produce high-quality depth map.
    }
    \label{fig:framework}
    \vspace{-4mm}
\end{figure*}

\section{Related works}
\label{sec:intro}

\subsection{Depth map super-resolution}
Depth map super-resolution (DMSR) has become an increasingly essential technique for enhancing the resolution of depth maps produced by various sensors \cite{hui2016depth, guo2018hierarchical, kiechle2013joint}. 
Typically, these sensors generate depth maps at a lower resolution compared to their corresponding RGB images, posing significant challenges for applications that require high-resolution depth maps \cite{zhong2023guided}. 
Hence, DMSR technology is critical for a wide range of applications, such as 3D reconstruction \cite{slavcheva2020variational}, virtual reality (VR) \cite{rokita1996generating}, and augmented reality (AR) \cite{zhou2020fine}, where high-quality depth maps are essential to ensure accuracy and immersive experiences.

Over the years, numerous methods have been proposed to tackle the challenges associated with depth map super-resolution (DMSR) \cite{song2020channel, he2021towards, xu2022depth, zhao2024joint, ye2020depth, dong2021learning, yuan2023recurrent, yan2022learning}. 
Traditional approaches for DMSR often rely on interpolation techniques such as bilinear or bicubic interpolation \cite{li2012joint}. 
While these methods are computationally efficient and straightforward to implement, they tend to produce depth maps that are overly smooth and lack important fine geometry details. 
Furthermore, these approaches fail to adequately preserve depth discontinuities, which leads to significant visual artifacts, particularly in regions with complex structures or sharp depth transitions \cite{zhong2023guided}. 
The shortcomings make them unsuitable for high-quality depth recovery, especially in scenarios requiring high accuracy and fine geometry details.

To overcome these limitations, convolutional neural networks (CNNs) have been widely employed to learn complex mappings from low-resolution (LR) to high-resolution (HR) depth maps \cite{hui2016depth, song2020channel}. 
CNN-based methods have demonstrated notable improvements in recovering fine geometry details and enhancing overall depth map quality \cite{kim2021deformable, wang2024scene, sun2021learning, yang2022codon, wang2024degradation}. 
More recently, transformer-based models have also been explored to capture long-range dependencies, leading to further advancements in DMSR \cite{liang2021swinir}.

\subsection{Guided depth map super-resolution}
Guided depth map super-resolution (GDSR) is designed to enhance the quality of low-resolution (LR) depth maps by leveraging corresponding high-resolution (HR) color images as guidance \cite{zhong2023guided}. 
By incorporating the additional information provided by color images, GDSR significantly enriches the quality of depth maps, adding fine geometry details and contextual elements that would otherwise be absent. 
A variety of approaches have been developed to tackle the challenges associated with GDSR, and these methods can broadly be categorized into filtering-based techniques and learning-based methods \cite{kou2015gradient, metzger2023guided, yuan2023structure}.

Filtering-based methods estimate depth by applying a weighted average to local pixels, where the weights are determined based on the similarity between pixels in RGB-D image pairs \cite{ham2017robust}. 
Popular techniques in this category include bilateral filtering \cite{jevnisek2017co}, non-local mean filtering \cite{huhle2010fusion}, and guided filtering \cite{kou2015gradient}. 
These methods are recognized for their computational efficiency and simplicity. 
However, they also come with certain limitations. 
When depth discontinuities do not align well with edges in the corresponding color image, artifacts are likely to occur. 
Additionally, their filter kernels are often designed for specific tasks, which reduces the adaptability and limits the flexibility \cite{zhong2023guided}.

Learning-based methods focus on leveraging the powerful feature extraction capabilities of convolutional neural networks (CNNs) to effectively enhance the resolution of depth maps \cite{zhao2022discrete}. 
These approaches utilize high-resolution RGB images as guidance to improve the resolution of depth maps, effectively transforming low-resolution inputs into high-resolution outputs. 
Specifically, these GDSR models exploit the structural similarities between depth maps and RGB images, which allows for more precise edge preservation and detailed reconstruction \cite{yuan2023structure}. 
Furthermore, advanced techniques such as attention mechanisms and feature fusion strategies have been introduced to refine the integration of RGB guidance and depth features, ultimately achieving superior performance \cite{yang2022codon}.

\section{Method}
We first present the overall framework of the proposed GDNet in section \ref{framework}. Next, we introduce two key components: fine geometry detail encoder and global geometry encoder in section \ref{module1} and \ref{module2}. Finally, the loss function is described in section \ref{loss}.
\subsection{Overall framework}
\label{framework}
In this paper, we propose GDNet for compressed depth map super-resolution to address the challenges introduced by bit-depth compression and random noise. 
Specifically, we introduce a decoupling strategy that separates the recovery of high-quality depth maps into two aspects: detailed geometry learning and global geometric feature extraction.
The overall framework of our GDNet is illustrated in Figure \ref{fig:framework}, which contains three main components: a fine geometry detail encoder, a global geometry encoder and a depth decoder.
It takes an image and a compressed depth map as inputs and generates a high-quality depth map, which can be formulated as follows:
\begin{equation}
	\hat{D_{hq}} = \text{GDNet}(I, D_{lq}),
\end{equation}
where $\text{GDNet}(\cdot)$ and $I$ denote the transformation of our GDNet and the corresponding RGB image, $D_{lq}$ and $\hat{D_{hq}}$ represent the compressed low-quality depth map and the recovered high-quality depth map, respectively.

\noindent\textbf{Fine geometry detail encoder.} To begin with, the input image is fed into the fine geometry detail encoder to extract the fine geometry details, which can be expressed as follows:
\begin{equation}
    F_{dg} = \text{FGDE}(I),
\end{equation}
where $\text{FGDE}(\cdot)$ and $F_{dg}$ denote the fine geometry detail encoder and detailed geometric features, respectively.

\noindent\textbf{Global geometry encoder.} The compressed depth map is processed by the global geometry encoder to extract global geometric features, which is illustrated as follows:
\begin{equation}
    F_{gg} = \text{GGE}(D_{lq}),
\end{equation}
where $\text{GGE}(\cdot)$ and $F_{gg}$ denote the global geometry encoder and global geometric features, respectively.

\noindent\textbf{Depth decoder.} Once we obtain detailed and global geometric features, a depth decoder is employed to reconstruct high-quality depth map:
\begin{equation}
    \hat{D_{hq}} = \text{DD}(F_{dg}, F_{gg}),
\end{equation}
where $\text{DD}(\cdot)$ represents the transformation of the depth decoder.
In the depth decoder, we first fuse detailed and global geometric features by using multilayer perceptrons (MLPs). 
Next, a depth prediction head equipped with depth bins centers is used to reconstruct high-quality depth map \cite{Agarwal_2023_WACV}.

\subsection{Fine geometry detail encoder}
\label{module1}
Bit-depth compression leads to uniform depth representations in regions with subtle variations, posing challenges for current models in accurately recovering intricate details. 
This loss of detail can blur critical features and diminish the overall quality of the depth map, making it harder to capture the detailed geometric information of the scene.

To tackle this challenge, we introduce fine geometry detail encoder (FGDE) that aims at maintaining the details in high-resolution low-level image features. 
These features are rich in texture and edge information, which are crucial for capturing detailed geometry within the scene. 
However, relying solely on these high-resolution low-level image features may not suffice to fully recover the detailed information required for accurate depth reconstruction.
To complement this, FGDE also incorporates complementary information from low-resolution context-level image features. 
By combining the strengths of both high-resolution low-level features and low-resolution context-level image features, FGDE facilitates effective extraction of detailed geometric information, thereby achieving better fine geometry detail recovery in regions with subtle depth variations. 

\begin{figure}[t]
    \centering
    \subfloat[The architecture of the proposed FGDE.]{%
    \includegraphics[width=0.49\textwidth]{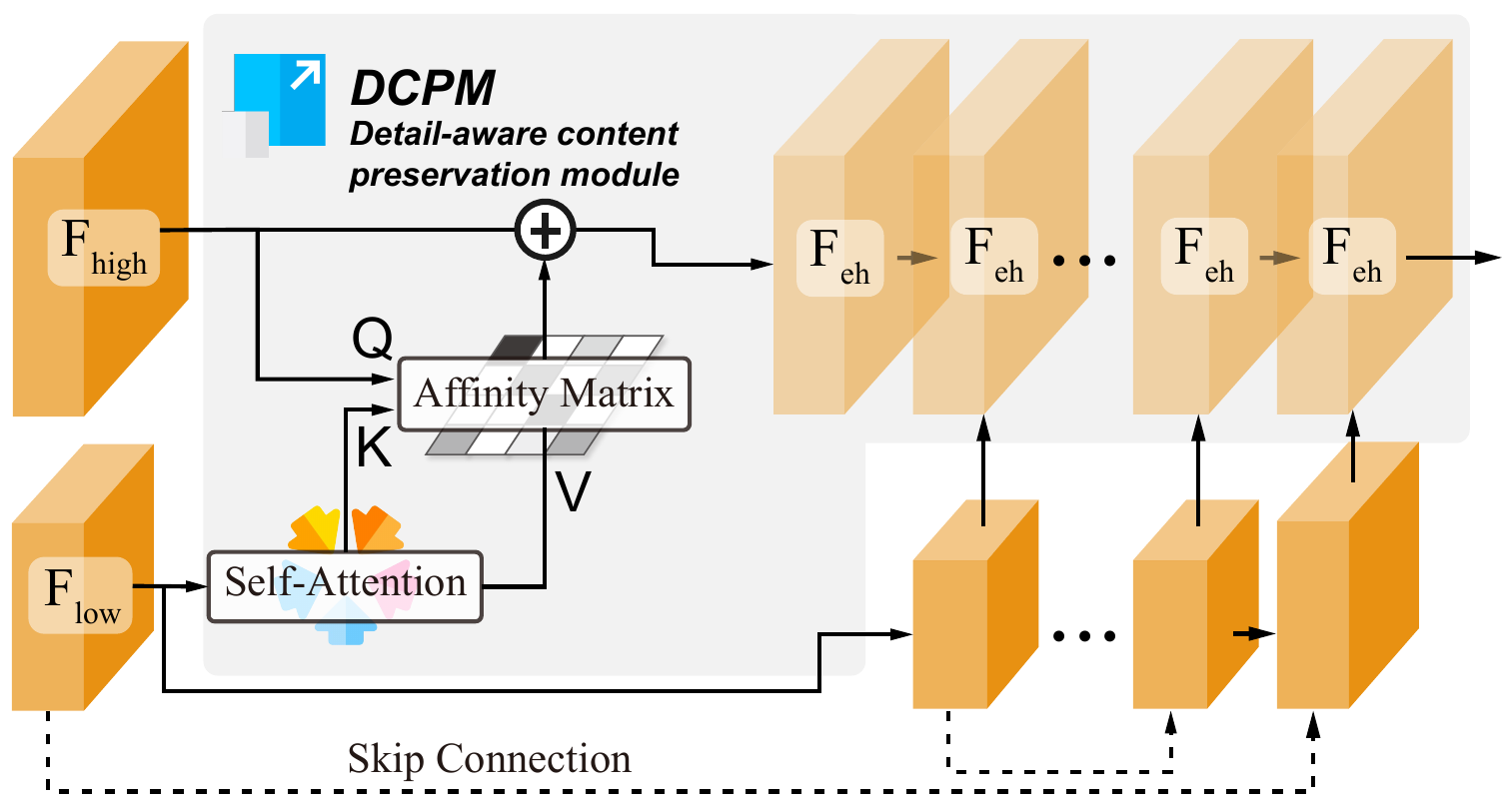}
    } \\
    \subfloat[The detailed process of the LFR in GGE.]{%
    \includegraphics[width=0.49\textwidth]{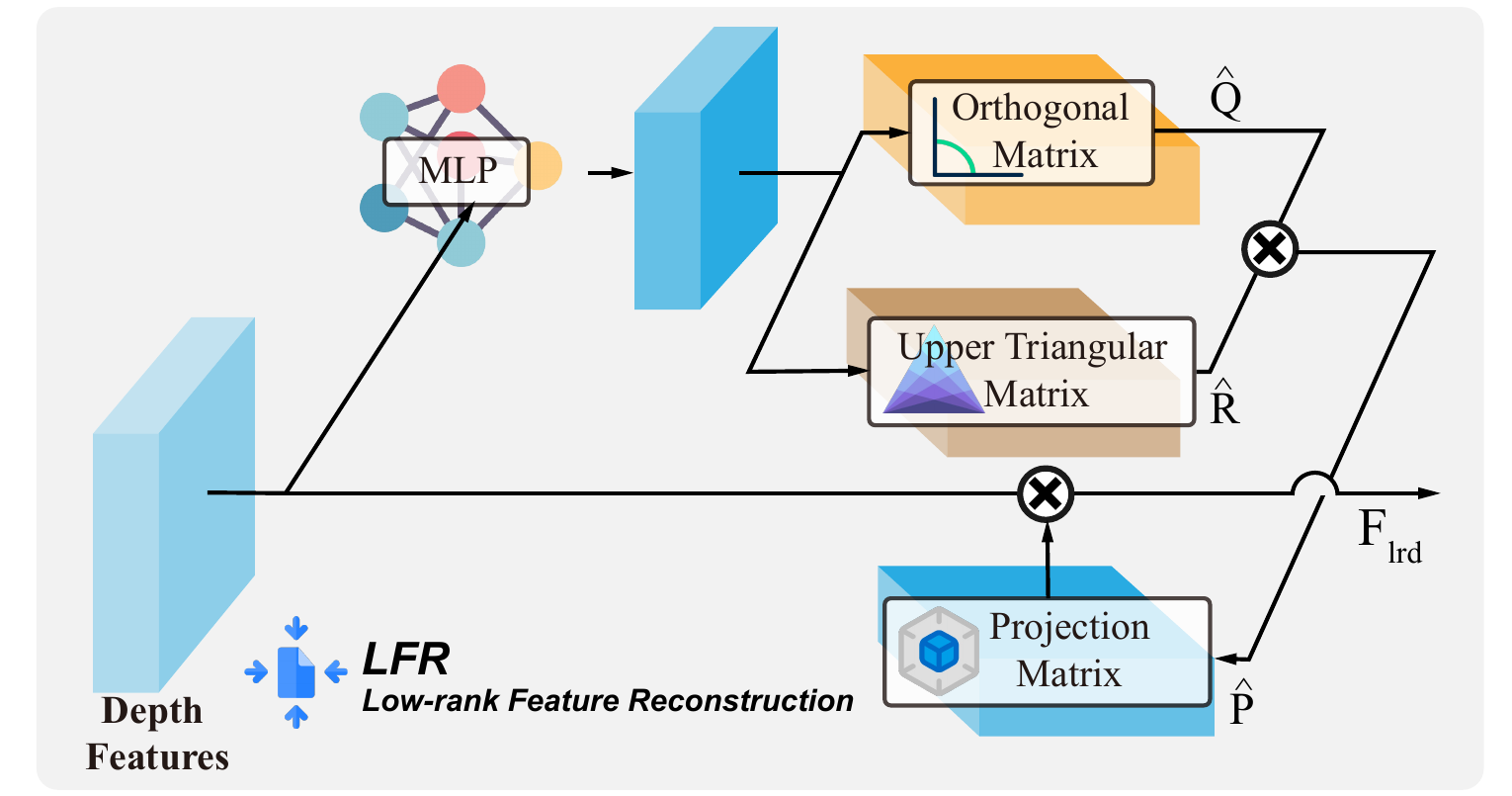}
    } \vspace{-2mm}
    \caption{\textbf{(a) The Detailed Structure of the Proposed Fine Geometry Detail Encoder (FGDE).} The purpose of FGDE is to preserve fine geometric details in high-resolution low-level image features while augmenting them with supplementary information derived from low-resolution context-level image features. \textbf{(b) The Process of Low-rank Feature Reconstruction in Global Geometry Encoder (GGE).} By integrating low-rank feature reconstruction, GGE aims at completing feature reconstruction in a low-rank space, thereby achieving the objectives of noise suppression and effective extraction of global geometric cues.
    }
    \label{fig:modules}
    \vspace{-6mm}
\end{figure}

The comprehensive architecture of the FGDE with detail-aware content preservation module (DCPM) is depicted in Figure \ref{fig:modules} (a). 
In detail, given low-resolution context-level image features, a self-attention module is employed to adaptively emphasize important spatial details while suppressing irrelevant information, enhancing the representation of fine textures. This process can be formulated by:
\begin{equation}
    F_{el} = \text{SA}(F_{low}),
\end{equation}
where $\text{SA}(\cdot)$, $F_{low}$ and $F_{el}$ denote the transformation of self attention, low-resolution context-level image features and the enhanced context-level image features, respectively. 

Next, we employ a cross-attention module to maximize the preservation of detailed content in the high-resolution low-level image features while incorporating complementary information from the low-resolution context-level image features.
In this module, the high-resolution low-level image features serve as the query, while the enhanced context-level image features act as the key and value. The detailed process can be expressed as:
\begin{equation}
    F_{eh} = \text{CA}(F_{high}, F_{el}, F_{el}),
\end{equation}
where $\text{CA}(\cdot)$, $F_{high}$ and $F_{eh}$ denote the transformation of cross-attention, high-resolution low-level image features and the output detail-aware features.
It is noted that in the fine geometry detail encoder, we employ DCPM several times to aggregate fine geometry details from high-resolution low-level image features and low-resolution context-level image features across different scales, as shown in Figure \ref{fig:framework}.

\subsection{Global geometry encoder}
\label{module2}
Densely distributed noise can significantly hinder the accurate estimation of global geometric structures in a scene, resulting in degraded depth map quality. 
To address this issue, we propose the global geometry encoder (GGE), designed to enhance both global geometry-aware feature representation and noise robustness.
By projecting the features into a low-rank space to obtain a compact representation, the GGE efficiently captures the underlying structural cues of the scene, filtering out noise while retaining essential global geometric information.

\noindent\textbf{Low-rank feature reconstruction}.
Given features $X \in \mathbb{R}^{n \times c}$, and low-rank basis vectors $B \in \mathbb{R}^{n \times d}, d < c$, the objective of low-rank feature reconstruction is to learn the representation in the low-rank space that preserves the global geometric information of the scene while suppressing the effects of noise. Specifically, the optimization goal for this reconstruction should be:
\begin{equation}
    \min_{R} \|X - BR\|_F^2,
    \label{equ:taget}
\end{equation} 
where $R$ denotes the reconstruction coefficient matrix in the low-rank space.
By solving Equation (\ref{equ:taget}), we can obtain the following results:
\begin{equation}
    R = (B^TB)^{-1}B^TX,
\end{equation}
where $(\cdot)^{-1}$ denotes the inverse of a matrix.
Hence, the low-rank projection matrix should be:
\begin{equation}
    P = B(B^TB)^{-1}B^T,
\end{equation}
where $P$ denotes the low-rank projection matrix.

However, it is difficult to directly learn a set of linearly independent vectors, which may lead to a problem: the inverse of \((B^T B)\) may not exist. To address this issue, we employ QR decomposition to construct a full-rank matrix, which serves as the basis vectors for the low-rank space. 
Specifically, we perform QR decomposition on $B$ as follows:
\begin{equation}
    Q, R = \text{QR}(B),
\end{equation}
where $\text{QR}(\cdot)$ denotes the transformation of QR decomposition.
Assuming the rank of \( B \) is \( r \), we denote the first \( r \) rows of \( Q \) as \( Q_r \) and the first \( r \) columns of \( R \) as \( R_r \). Consequently, we can construct the following full-rank matrix:
\begin{equation}
    \hat{B} = Q_rR_r,
\end{equation}
where $\hat{B}$ can be regarded as the basis vectors of low-rank space.
Hence, the low-rank projection matrix should be:
\begin{equation}
    \begin{split}
        P &= \hat{B}(\hat{B}^T\hat{B})^{-1}\hat{B}^T \\
          &= Q_rR_r((Q_rR_r)^TQ_rR_r)^{-1}(Q_rR_r)^T.
    \end{split}
    \label{equ:pro}
\end{equation}
\captionsetup[subfloat]{labelsep=none,format=plain,labelformat=empty}
\begin{figure*}[t]
    \centering
    \subfloat[RGB Image]{%
        \includegraphics[width=0.245\textwidth]{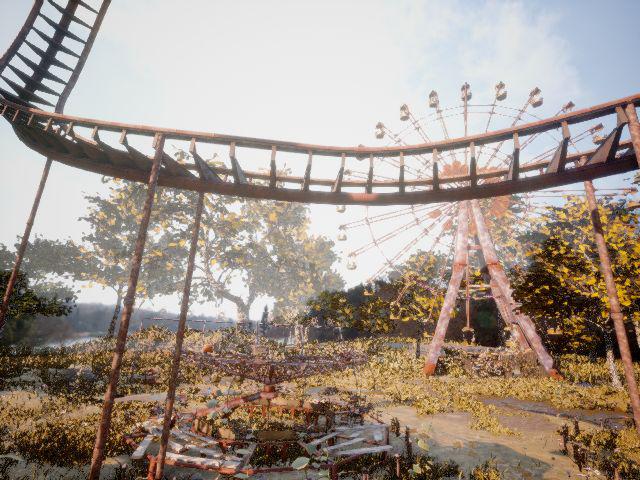}
    }
    \subfloat[GT Depth Map]{%
        \includegraphics[width=0.245\textwidth]{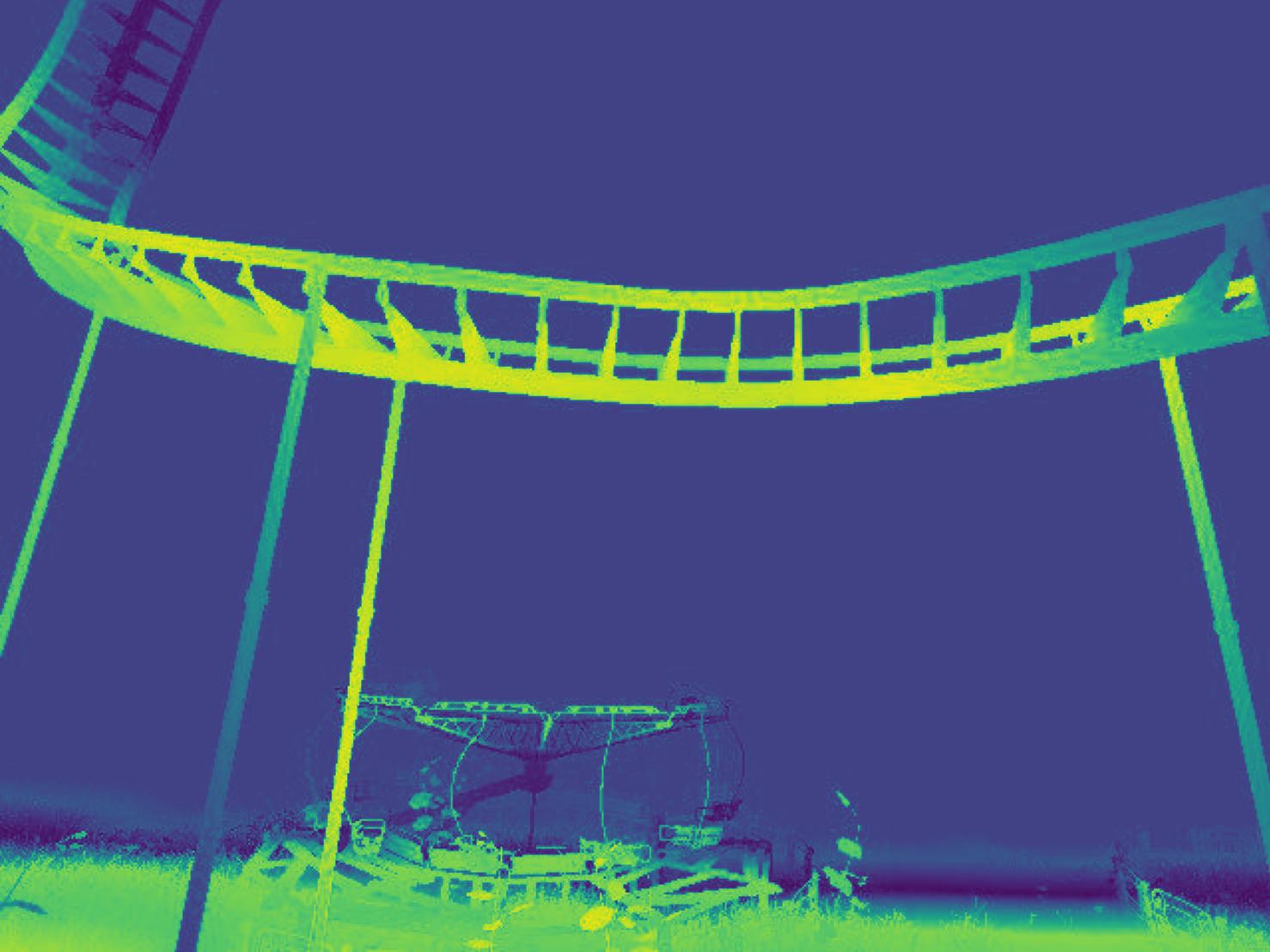}
    }
    \subfloat[Compressed Depth Map]{%
        \includegraphics[width=0.245\textwidth]{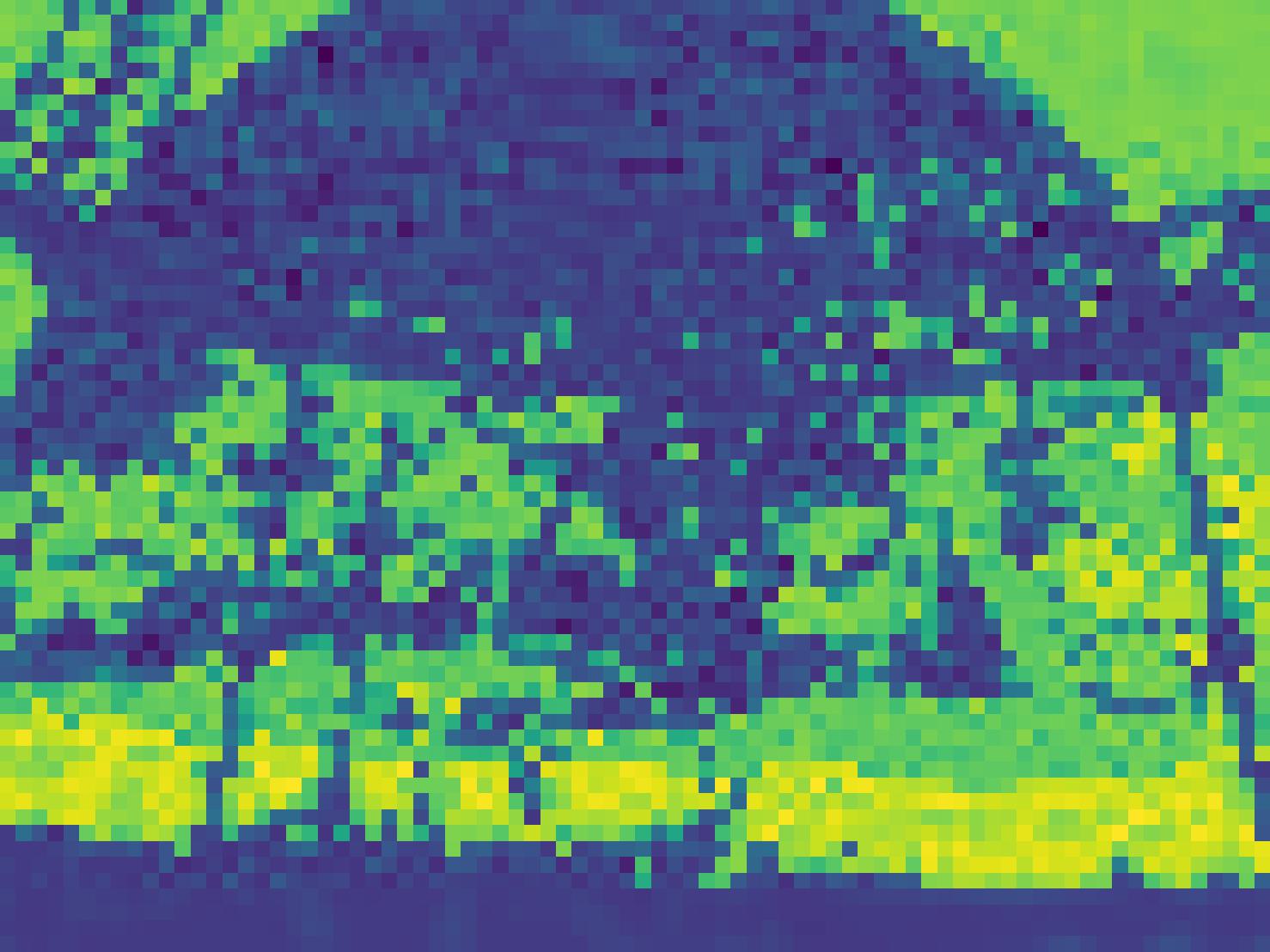}
    }
    \subfloat[DAv2 ++]{%
        \includegraphics[width=0.245\textwidth]{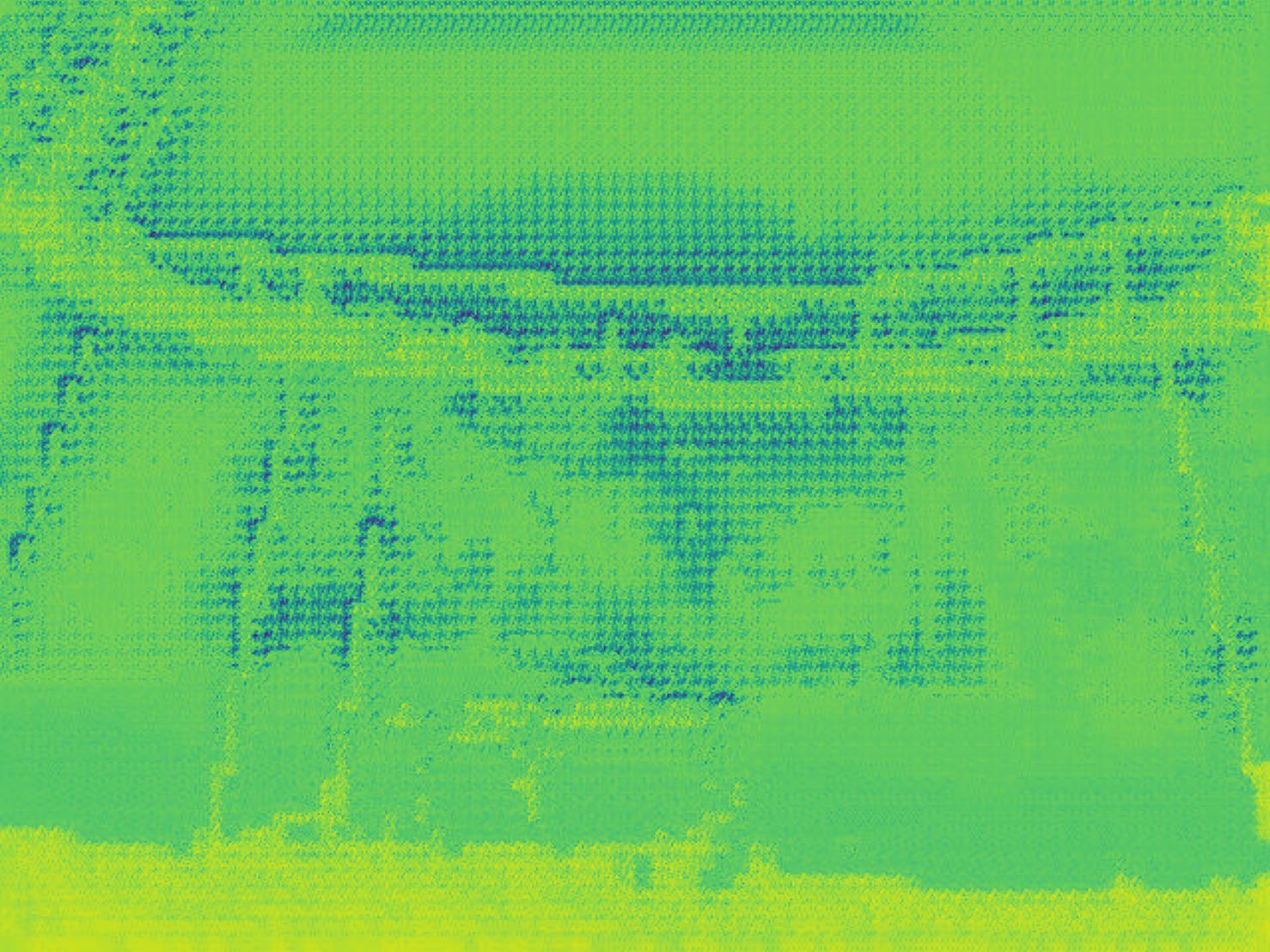}
    }\\
    \subfloat[DAS-Depth]{%
        \includegraphics[width=0.245\textwidth]{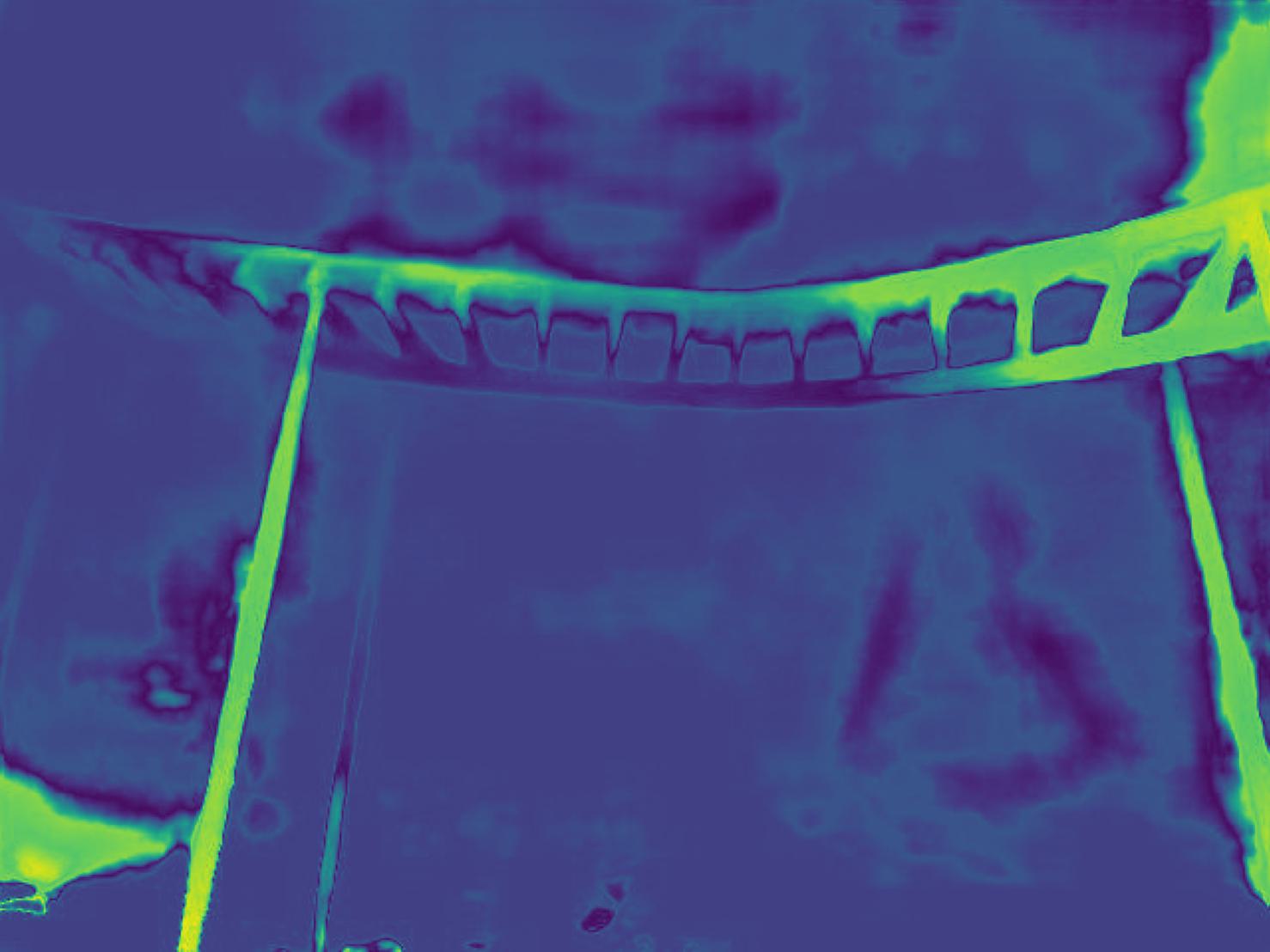}
    } 
    \subfloat[RAFT-DU]{%
        \includegraphics[width=0.245\textwidth]{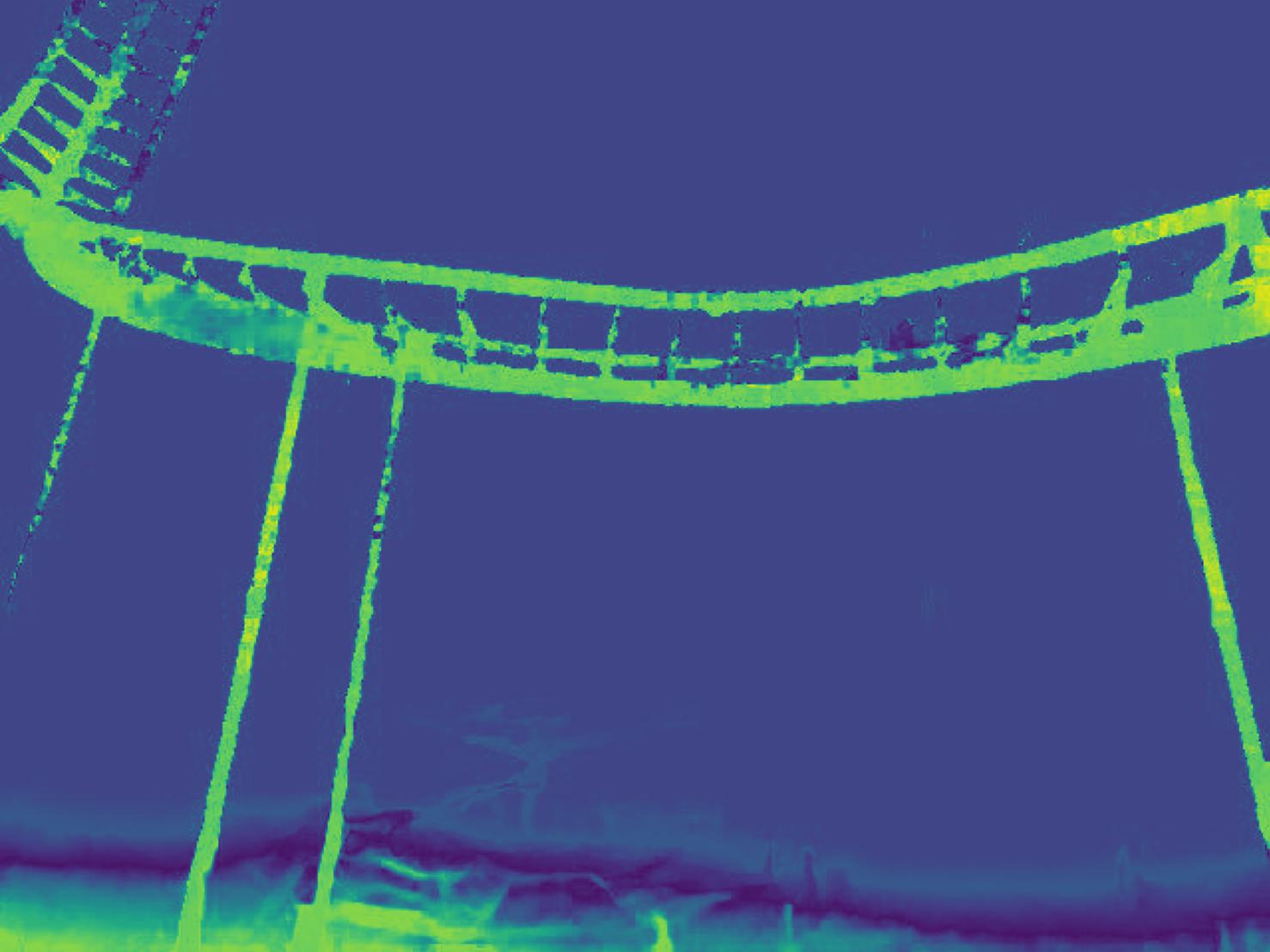}
    } 
    \subfloat[DINOv2-ControlNet]{%
        \includegraphics[width=0.245\textwidth]{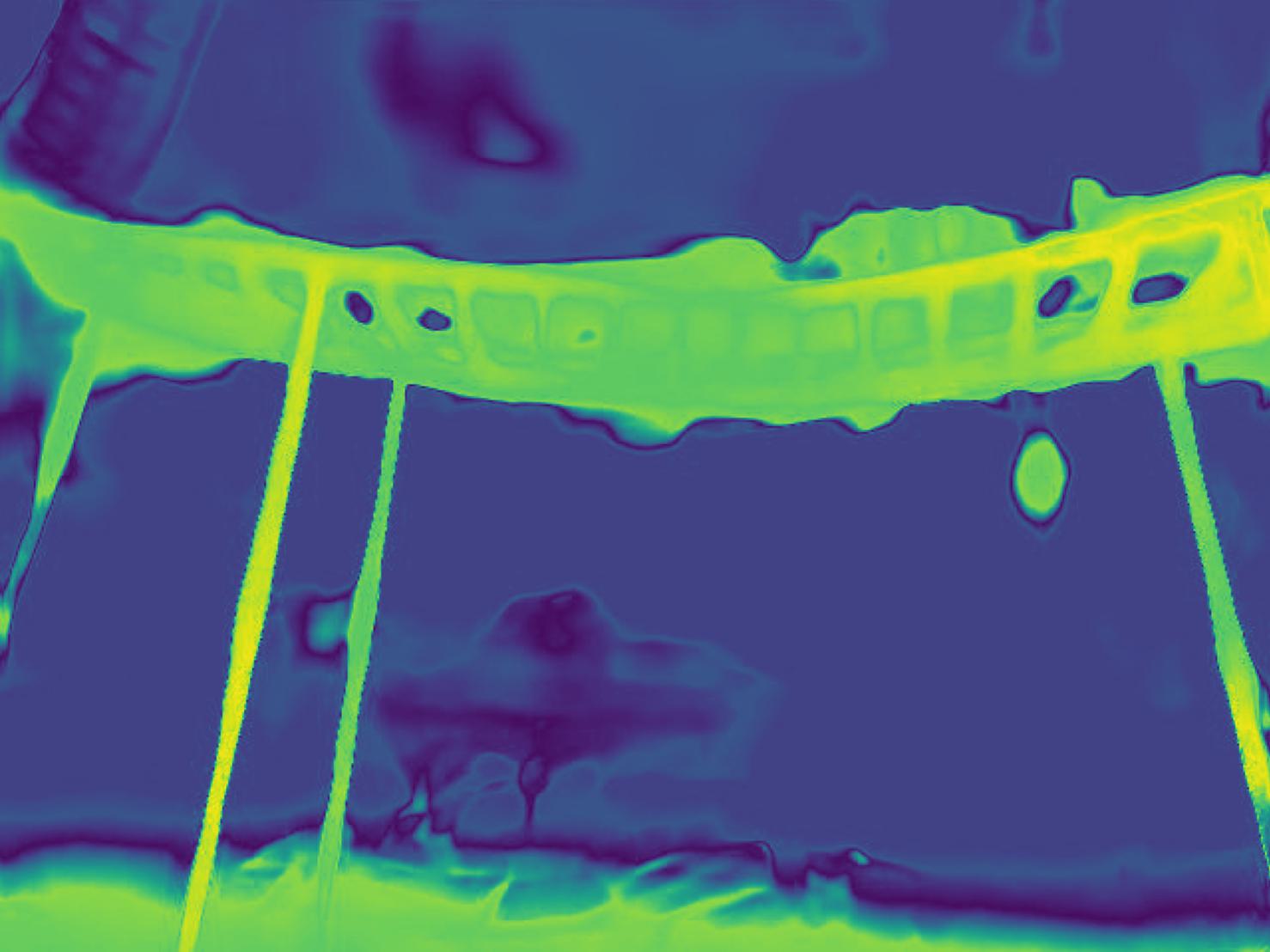}
    }
    \subfloat[Ours]{%
        \includegraphics[width=0.245\textwidth]{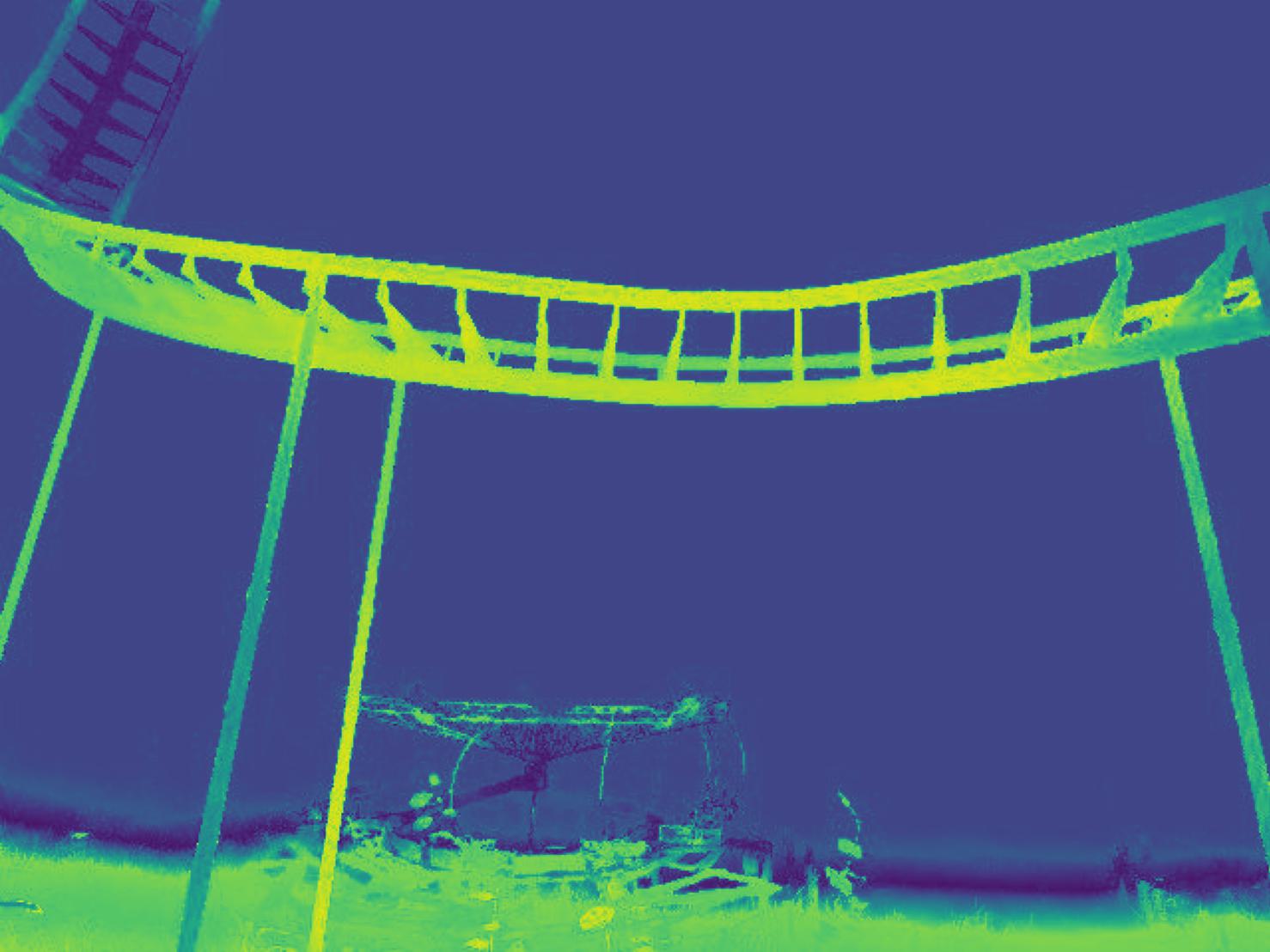}
    }\\
    \subfloat[Error Map of DAS-Depth]{%
        \includegraphics[width=0.245\textwidth]{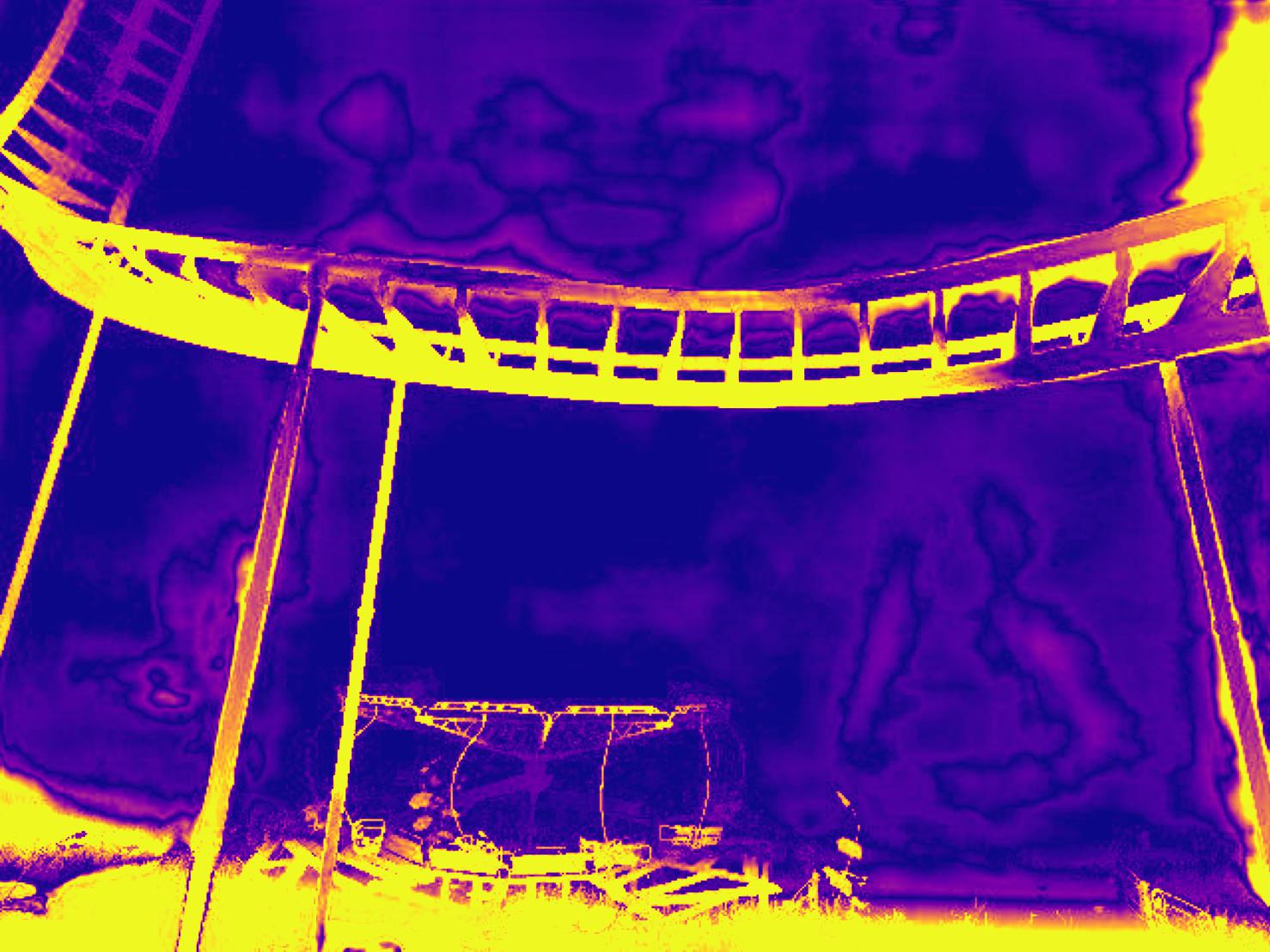}
    } 
    \subfloat[Error Map of RAFT-DU]{%
        \includegraphics[width=0.245\textwidth]{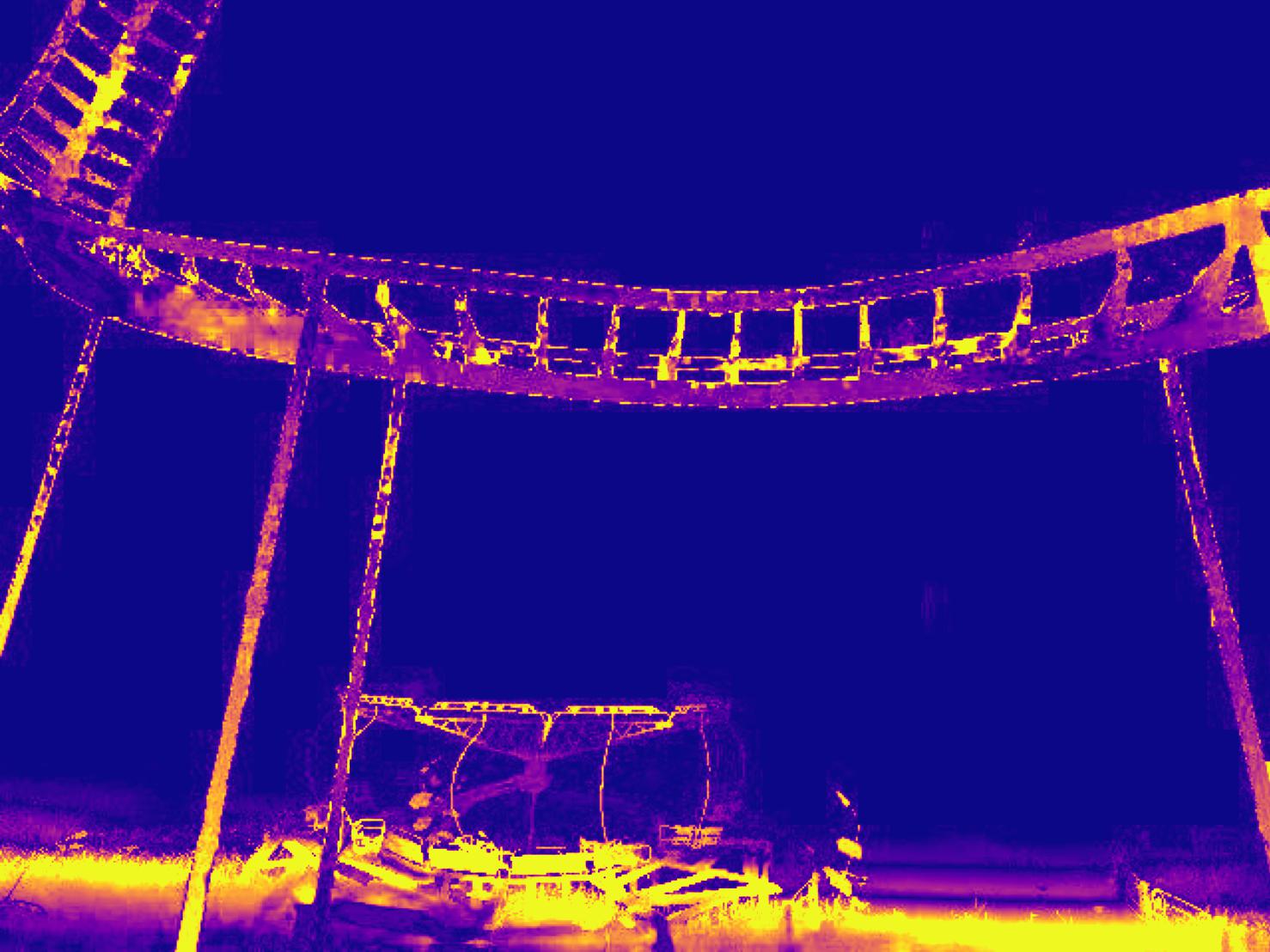}
    } 
    \subfloat[Error Map of DINOv2-ControlNet]{%
        \includegraphics[width=0.245\textwidth]{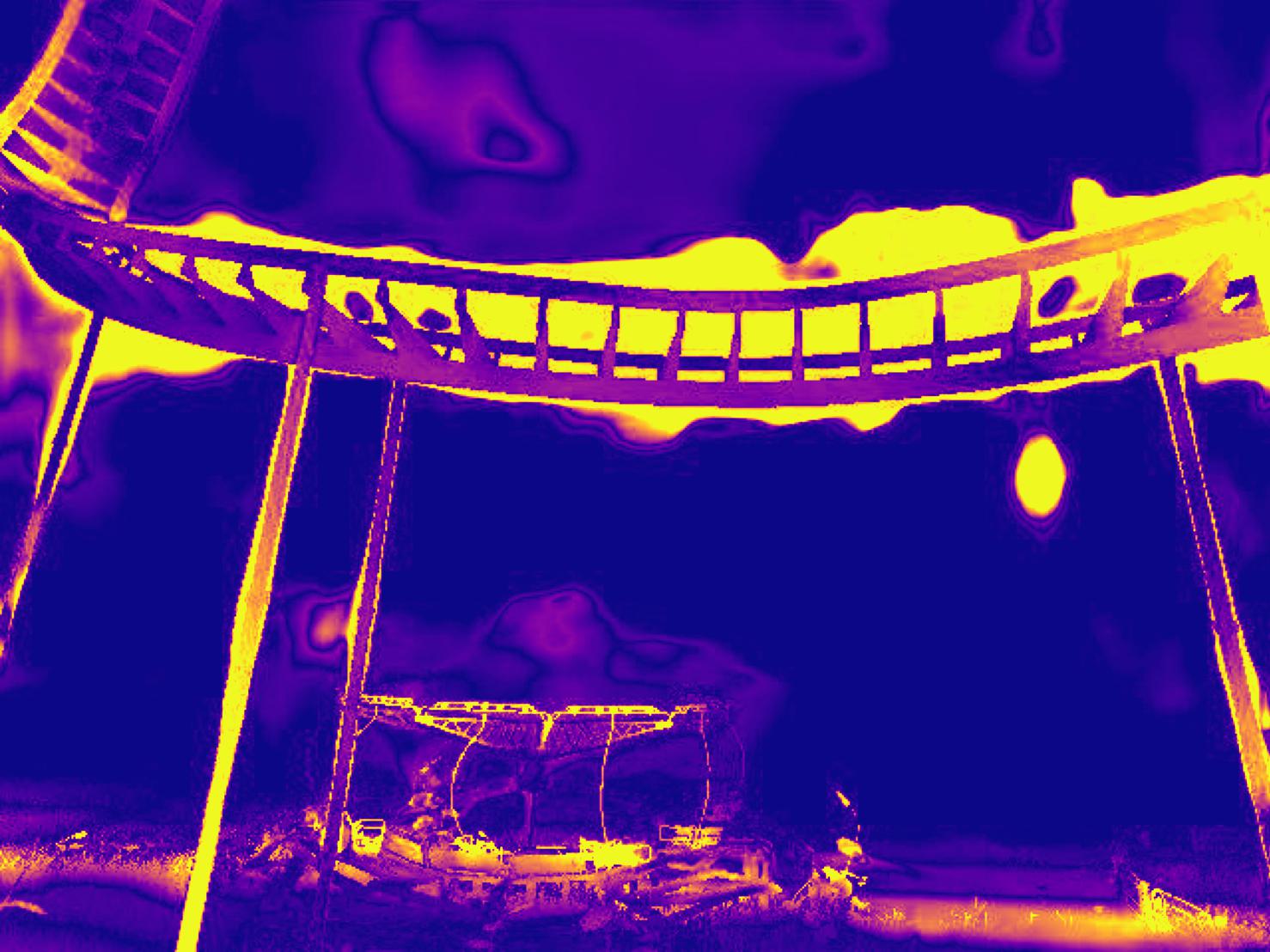}
    } 
    \subfloat[Error Map of Ours]{%
        \includegraphics[width=0.245\textwidth]{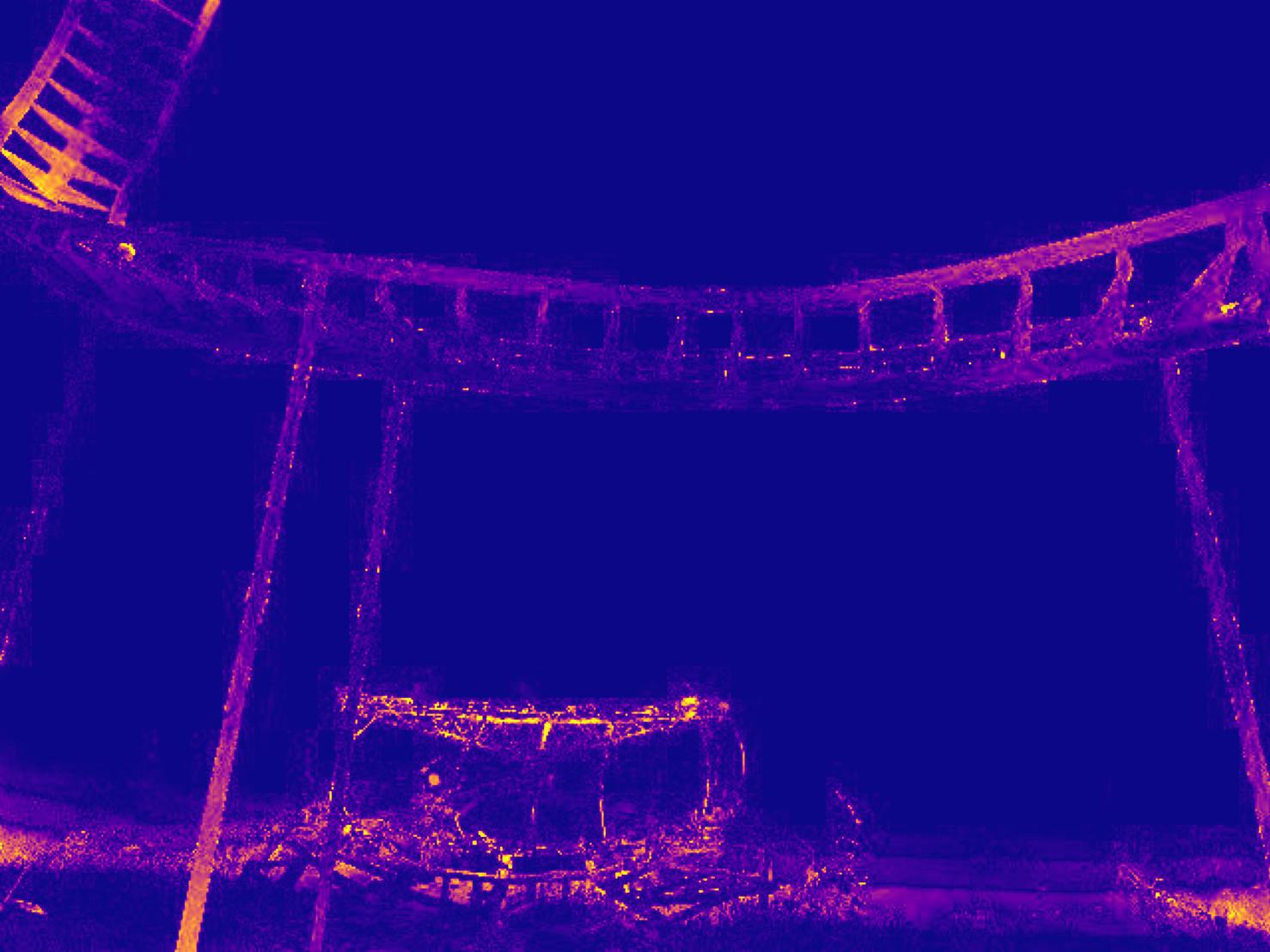}
    }\\
    \vspace{-2mm}
    \caption{\textbf{Visual Comparisons on AIM 2024 Compressed Depth Upsampling Challenge Dataset. }
    The first two rows show the RGB image, ground truth depth map, compressed depth map, and predicted depth maps produced by various methods. The third row displays the error maps for each method. In the error map, darker areas indicate smaller errors.
    Our method shows superior performance in recovering details within regions of subtle depth variation, offering more accurate predictions of the geometric structure. Furthermore, error maps reveal that the depth errors in our predictions are notably lower than those produced by other methods.}
    \label{fig:aim}
    \vspace{-4mm}
\end{figure*}

\noindent\textbf{Full pipeline.}
We introduce the GGE, which is designed to derive a robust low-rank representation of depth features. 
This module is specifically tailored to enhance feature representation by filtering out noise while preserving essential global geometric information. 
Figure \ref{fig:modules} (b) shows the detailed process of low-rank feature reconstruction (LFR) in the proposed GGE.
LFR aims at reconstructing compact low-rank features, which can be further divided into two core components: low-rank basis vector learning and low-rank projection. 
In the first step, low-rank basis vector learning aims to obtain a set of linearly independent vectors as basis vectors.
Next, low-rank projection projects depth features into a low-rank space, enabling the reconstruction of noise-suppressed and global geometry-retained features. 

\noindent\textbf{Low-rank basis vectors learning}.
During the low-rank basis vector learning process, we begin by using a MLP, generating a set of low-rank vectors that capture the implicit correlations between depth features and the low-rank space. This process can be expressed as:
\begin{equation}
    F_{lv} = \text{MLP}(F_{depth}),
\end{equation}
where $F_{lv}$ and $\text{MLP}(\cdot)$ denote the initialized low-rank vectors and the transformation of the MLP.

Next, we perform QR decomposition on the initialized low-rank vectors \(F_{lv}\):
\begin{equation}
    \hat{Q}, \hat{R} = \text{QR}(F_{lv}).
\end{equation}
Consequently, we can obtain the full-rank matrix $\hat{Q}_r\hat{R}_r$, which is treated as the basis vectors in the low-rank space.

\noindent\textbf{Low-rank projection}.
Once the low-rank basis vectors are obtained, the low-rank projection matrix can be derived using Equation (\ref{equ:pro}), which can be described by:
\begin{equation}
    \hat{P} = \hat{Q}_r\hat{R}_r((\hat{Q}_r\hat{R}_r)^T\hat{Q}_r\hat{R}_r)^{-1}(\hat{Q}_r\hat{R}_r)^T,
\end{equation}
where $\hat{P}$ denotes the corresponding low-rank projection matrix. 
Notably, due to the inherent numerical instability of matrix inversion, we utilize the Neumann series \cite{liu2024neumann} as an effective approximation for the matrix inverse.

Finally, we apply the low-rank projection matrix $\hat{P}$ to project the depth features into the low-rank space, yielding the low-rank feature representation:
\begin{equation}
    F_{lrd} = \hat{P}F_{depth},
\end{equation}
where $F_{lrd}$ denotes the reconstructed low-rank depth features.
The compact representation $F_{lrd}$ not only preserves the global geometric information of the scene but also effectively mitigates the adverse effects of noise, ensuring more robust and accurate representation in the low-rank space.

\subsection{Loss function}
\label{loss}
We follow the previous work for depth estimation \cite{Agarwal_2023_WACV}, adopting the Scale Invariant (SILog) loss as the objective function for training GDNet. 
The SILog loss emphasizes relative depth differences and enhances the network's sensitivity to subtle variations, leading to more accurate and perceptually consistent depth reconstructions.
The details of SILog loss can be expressed as follows:
\begin{equation}
    G_{i}=\log(\hat{D_{hq}^{i}}) - \log(D_{hq}^{i}),
\end{equation}
\begin{equation}
    \mathcal{L}_{SILog} = \alpha \sqrt{\frac{1}{n} \sum_{i} G_{i}^{2} - \frac{\lambda}{n^{2}}\left(\sum_{i} G_{i}\right)^{2}},
\end{equation}
where $\hat{D_{hq}^{i}}$ and $D_{hq}^{i}$ denote the predicted high-quality depth map and ground-truth depth map at pixel $i$, respectively, and $n$ represents the total number of pixels in an image. Additionally, $\lambda$ and $\alpha$ are hyperparameters which are set to 0.85 and 10 in our experiments, respectively.

\begin{figure*}[t]
    \centering
    \subfloat[GT Depth Map]{%
        \includegraphics[width=0.245\textwidth]{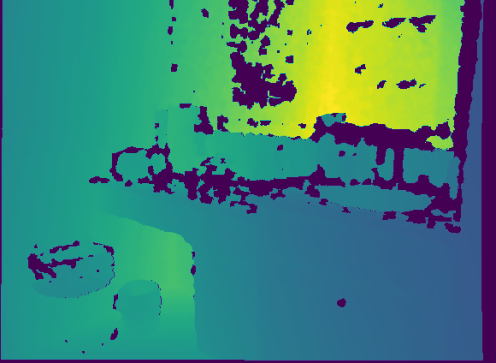}
    }
    \subfloat[Depth Anything~\cite{depth_anything_v1}]{%
        \includegraphics[width=0.245\textwidth]{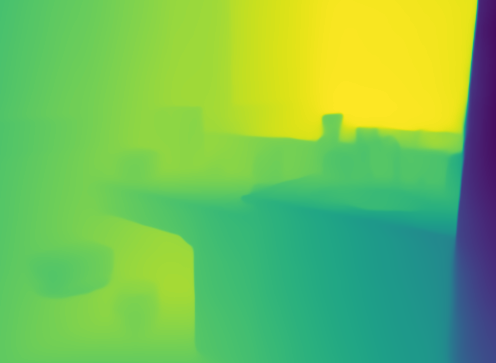}
    } 
    \subfloat[SGNet~\cite{wang2024sgnet} ]{%
        \includegraphics[width=0.245\textwidth]{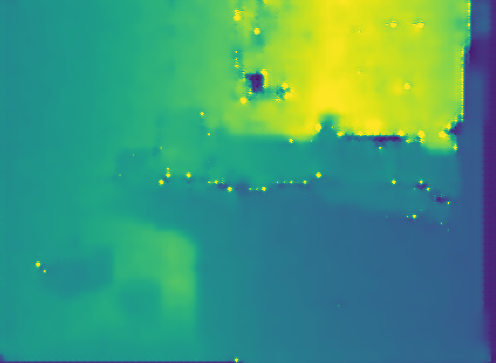}
    }
    \subfloat[Ours]{%
        \includegraphics[width=0.245\textwidth]{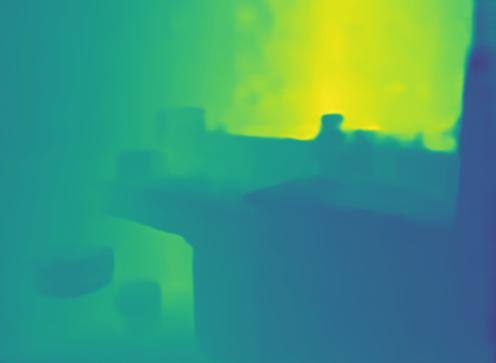}
    } \\
    \subfloat[RGB Image]{%
        \includegraphics[width=0.245\textwidth]{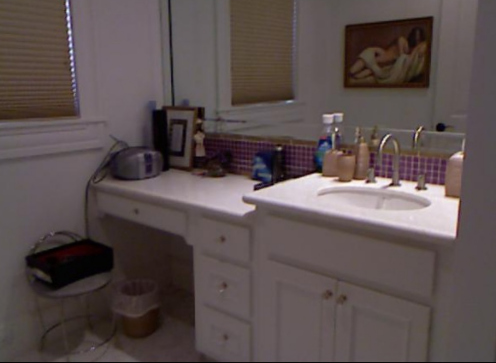}
    }
    \subfloat[Error Map of Depth Anything]{%
        \includegraphics[width=0.245\textwidth]{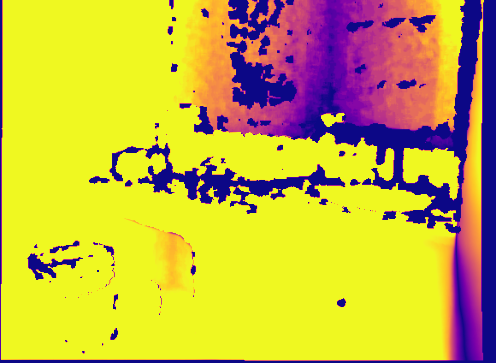}
    } 
    \subfloat[Error Map of SGNet]{%
        \includegraphics[width=0.245\textwidth]{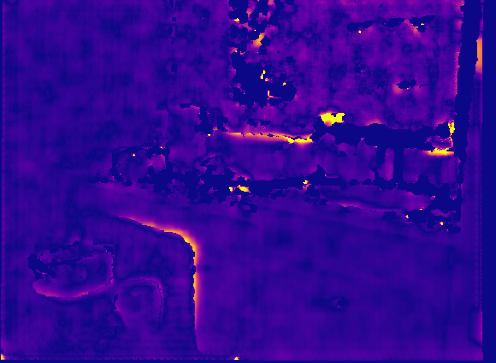}
    }
    \subfloat[Error Map of Ours]{%
        \includegraphics[width=0.245\textwidth]{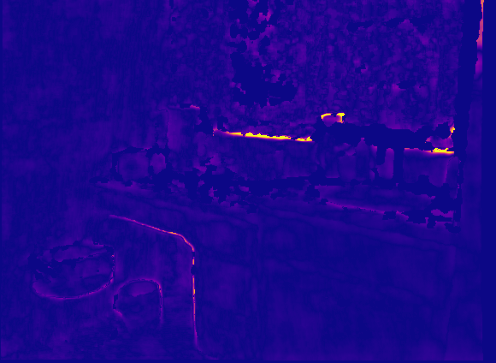}
    }
    \vspace{-2mm}
    \caption{\textbf{Visual Comparisons on Compressed-NYU Dataset between Our GDNet and Other Well-known Methods.} The first row displays the ground truth depth alongside the predicted results of various approaches, while the second row shows the corresponding RGB image and error maps. For error maps, darker areas indicate smaller errors. As demonstrated, our method generates depth maps that are highly consistent with the ground truth and exhibit significantly reduced errors compared to other techniques. In regions with fine geometry details, especially along the edges, other methods often produce blurred results, while our approach delivers much sharper outputs.
    }
    \label{fig:nyu}
    \vspace{-4mm}
\end{figure*}

\section{Experiments}
\subsection{Experimental settings}
\textbf{Evaluated datasets.} 
In our experiments, we utilize two datasets to validate the effectiveness of our GDNet. 
First, we employ the dataset proposed in \cite{conde2024compressed}, which is derived from the TartanAir dataset \cite{wang2020tartanair} and used in AIM 2024 Compressed Depth Upsampling Challenge. 
This dataset includes a subset of RGB images and depth maps from various scenes, with 3,866 samples used for training and 257 samples for testing. 
All compressed depth maps in this dataset undergo simultaneous bit-depth compression, downsampling, and random noise.
Additionally, we synthesize a new dataset termed Compressed-NYU based on the NYU Depth Dataset V2 \cite{silberman2012indoor}, following the approach in \cite{conde2024compressed}. 
Among the data, 795 samples were used for training, and 654 samples were used for testing.

\noindent\textbf{Implementation details.} 
All experiments are performed using the PyTorch framework in a Python environment, utilizing one NVIDIA A100 GPU. During training, we apply random cropping, as well as random horizontal and vertical flipping, as part of our data augmentation strategy. The Adam optimizer is used with a batch size of 2.
We train GDNet over 400 epochs, initializing the learning rate at 2e-4 and gradually reducing it linearly to 5e-6. 
To assess the effectiveness of our method for compressed depth map super-resolution, we employ two standard metrics: Mean Absolute Error (MAE) and Root Mean Square Error (RMSE).
In particular, lower MAE and RMSE values signify higher quality of the reconstructed depth map, indicating a closer match to the ground-truth.
To give a thorough comparison, we include several well-known methods, such as SGNet \cite{wang2024sgnet}, Depth Anything \cite{depth_anything_v1} and Depth Anything V2 \cite{depth_anything_v2}. 

\begin{table}[t]
    \caption{\textbf{Quantitative Performance Comparison of Various Methods on the AIM 2024 Compressed Depth Upsampling Challenge Dataset.} The table reports both MAE and RMSE. 
    Our proposed method outperforms the existing approaches, achieving the lowest MAE and RMSE, indicating superior accuracy in depth map super-resolution under compression scenarios.
    }
    \vspace{-2mm}
    \label{tab:aim}
    \centering
    \small
    \setlength{\tabcolsep}{16pt}
    \resizebox{\linewidth}{!}{
    \begin{tabular}{l|cccccc}
    \toprule[1pt]
    \rowcolor{gray!20}
    Method & MAE $\downarrow$ & RMSE $\downarrow$ \\
    \midrule[0.75pt]
    Bicubic & 16.480 & 57.690 \\
    DAS-Depth \cite{conde2024compressed} & \underline{0.294} & \underline{0.432} \\
    DINOv2-ControlNet \cite{conde2024compressed} & 0.498 & 0.816 \\
    RGA Inc. \cite{conde2024compressed} & 0.512 & 1.621 \\
    RAFT-DU \cite{conde2024compressed} & 1.506 & 2.935 \\
    DAv2 ++ \cite{conde2024compressed} & 1.939 & 2.140 \\
    SGNet \cite{wang2024sgnet} & 1.337 & 1.854 \\
    Depth Anything V2 \cite{depth_anything_v2} & 2.193 & 2.388 \\
    \midrule[0.75pt]
    \rowcolor{gray!20}
    Ours & \textbf{0.212} & \textbf{0.375} \\
    \bottomrule[1pt]
    \end{tabular}}
    \vspace{-4mm}
\end{table}

\begin{table}[t]
    \caption{\textbf{Quantitative Performance Comparison on the Synthesized Compressed-NYU Dataset}, evaluating different methods using both MAE and RMSE. 
    The results demonstrate the effectiveness of our approach, which achieves the best performance with significantly lower error metrics compared to other methods.
    }
    \vspace{-2mm}
    \label{tab:nyu}
    \centering
    \small
    \setlength{\tabcolsep}{10pt}
    \resizebox{\linewidth}{!}{
    \begin{tabular}{l|cccccc}
    \toprule[1pt]
    \rowcolor{gray!20}
    Method & MAE $\downarrow$ & RMSE $\downarrow$ \\
    \midrule[0.75pt]
    Depth Anything \cite{depth_anything_v1} & 1.1360 & 1.2525 \\
    Depth Anything V2-Small \cite{depth_anything_v2} & 0.5130 & 0.6026 \\
    Depth Anything V2-Base \cite{depth_anything_v2} & 0.5237 & 0.6037 \\
    Depth Anything V2-Large \cite{depth_anything_v2} & 0.5247 & 0.5994  \\
    UniDepth V1 \cite{piccinelli2024unidepth} & 0.1702 & 0.2404 \\
    UniDepth V2 \cite{piccinelli2024unidepth} &  0.1414 & 0.1944 \\
    iDisc \cite{piccinelli2023idisc} & 0.2735 & 0.3778 \\
    AdaBins \cite{bhat2021adabins} & 0.4233 & 0.5694 \\
    SGNet \cite{wang2024sgnet} & \underline{0.0426} & \underline{0.1051} \\
    \midrule[0.75pt]
    \rowcolor{gray!20}
    Ours & \textbf{0.0322} & \textbf{0.0739} \\
    \bottomrule[1pt]
    \end{tabular}}
    \vspace{-4mm}
\end{table}

\subsection{Quantitative results}
The experimental results, as summarized in Tables \ref{tab:aim} and \ref{tab:nyu}, demonstrate the superior performance of our proposed method compared to several state-of-the-art approaches on both the AIM 2024 Compressed Depth Upsampling Challenge dataset and the synthesized Compressed-NYU dataset.
On the  AIM 2024 Compressed Depth Upsampling Challenge dataset, our method achieves a MAE of 0.212 and an RMSE of 0.375, significantly outperforming existing methods such as DAS-Depth (MAE: 0.294, RMSE: 0.432) and DINOv2-ControlNet (MAE: 0.498, RMSE: 0.816). 
Notably, our approach also delivers a remarkable improvement over simpler interpolation-based methods like Bicubic (MAE: 16.48, RMSE: 57.69). 
Similarly, on the synthesized Compressed-NYU dataset, our method achieves the lowest MAE of 0.0322 and RMSE of 0.739, which is a substantial improvement compared to SGNet, the next best-performing method, with a MAE of 0.0426 and an RMSE of 0.1051. 
These results clearly demonstrate that our method is highly effective in depth map super-resolution under compression scenarios, showing its superiority in preserving fine geometry details and maintaining global geometric consistency.

\subsection{Visual results}
The visual comparisons, as illustrated in Figures \ref{fig:aim} and \ref{fig:nyu}, offer further comprehensive qualitative evidence supporting the effectiveness of our proposed method in producing high-quality depth maps.
In Figure \ref{fig:aim}, we present visual comparisons on the AIM 2024 Compressed Depth Upsampling Challenge dataset between our GDNet and other state-of-the-art methods. 
It is evident that our method more effectively recovers the depth information of the roller coaster track and ground facilities. 
The error map further highlights that the depth map reconstructed by our method exhibits significantly lower error compared to other approaches.
Additionally, Figure \ref{fig:nyu} provides visual comparisons on the Compressed-NYU dataset between our GDNet and competing methods. 
Other methods struggle to accurately process edge depth information and perceive the depth of certain objects within the scene, often resulting in noticeable blurring.
In contrast, our method demonstrates a superior capability in preserving and recovering detailed edge information.
Overall, these results demonstrate that our method produces higher-quality depth maps, with enhanced recovery of fine geometry details and better geometric consistency.

\begin{table}[t]
    \caption{\textbf{An Ablation Study of Our GDNet Conducted on the Synthesized Compressed-NYU Dataset.} In this study, we assess the contributions of each module and loss function within GDNet.}
    \vspace{-2mm}
    \label{tab:ablation}
    \centering
    \small
    \setlength{\tabcolsep}{4pt}
    \resizebox{\linewidth}{!}{
    \begin{tabular}{ccccccc|cc}
    \toprule[1pt]
    \rowcolor{gray!20}
    FGDE & DCPM & GGE & LFR& L1 & MSE & SILog & MAE $\downarrow$ & RMSE $\downarrow$ \\
    \midrule[0.75pt]
    & & \checkmark & \checkmark & & & \checkmark  & 0.0488 & 0.1079 \\
    \checkmark & & \checkmark & \checkmark & & & \checkmark  & 0.0475  & 0.0974 \\
    \checkmark & \checkmark & & & & & \checkmark & 0.3781 & 0.4949  \\
    \checkmark & \checkmark & \checkmark & & & & \checkmark & 0.0566 & 0.1048  \\
    \checkmark & \checkmark & \checkmark & \checkmark & \checkmark & & & 0.0347 & 0.0796  \\
     \checkmark & \checkmark & \checkmark & \checkmark & & \checkmark & & 0.0372 & 0.0796  \\
    \midrule[0.75pt]
    \rowcolor{gray!20}
    \checkmark & \checkmark & \checkmark & \checkmark & & & \checkmark & \textbf{0.0322} & \textbf{0.0739}  \\
    \bottomrule[1pt]
    \end{tabular}}
    \vspace{-4mm}
\end{table}

\subsection{Ablation study}
To evaluate the effectiveness of each critical component in GDNet, we perform a comprehensive ablation studies.

\noindent\textbf{Effectiveness of FGDE.}
To demonstrate the effectiveness of the proposed FGDE, we remove it from the complete pipeline.
Table \ref{tab:ablation} presents the performance impact of excluding FGDE from the full pipeline. 
A noticeable decline in performance is observed when FGDE is omitted, attributed to FGDE’s capacity to aggregate fine geometry details within the image, which is essential for reconstructing high-quality depth maps.

\noindent\textbf{Effectiveness of GGE.}
We further investigated the impact of the GGE for compressed depth map super-resolution.
As shown in Table \ref{tab:ablation}, both MAE and RMSE values increase in the absence of the GGE, reflecting a notable reduction in the quality of the reconstructed depth maps. 
This decline occurs because the proposed GGE performs feature reconstruction within a low-rank space, which not only mitigates the influence of noise but also facilitates the extraction of global geometric information, thereby enhancing the global geometric consistency of the reconstructed depth maps.

\noindent\textbf{Effectiveness of SILog loss.}
L1 loss and MSE loss are widely adopted for image and depth map super-resolution \cite{wang2024sgnet}. 
In our approach, we utilize SILog loss specifically for compressed depth map super-resolution.
To evaluate the effectiveness of SILog loss, we trained our GDNet using either L1 loss or MSE loss.
As presented in Table \ref{tab:ablation}, the numerical results indicate that training with SILog loss yields superior performance compared to other loss functions, underscoring its effectiveness in our framework.

\begin{table}[t]
    \caption{\textbf{An Ablation Study of the Number of CA and SA.} $N_{SA}$ and $N_{CA}$ denotes the numbers of SA and CA, respectively.}
    \vspace{-2mm}
    \label{tab:nyu_per}
    \centering
    \small
    \setlength{\tabcolsep}{14pt}
    \renewcommand{\arraystretch}{0.95}
    \resizebox{\linewidth}{!}{
    \begin{tabular}{l|cccccc}
    \toprule[1pt]
    \rowcolor{gray!20}
    Method & $N_{SA}$ & $N_{CA}$ & MAE $\downarrow$ & RMSE $\downarrow$ \\
    \midrule[0.75pt]
    $\text{S}_2\text{C}_1$ & 2 & 1 & 0.0357 & 0.0789  \\
    $\text{S}_3\text{C}_1$ & 3 & 1 & 0.0385  & 0.0829 \\
    $\text{S}_1\text{C}_2$ & 1 & 2 & 0.0345 & 0.0765 \\
    $\text{S}_1\text{C}_3$ & 1 & 3 & 0.0372 & 0.0796  \\
    \midrule[0.75pt]
    \rowcolor{gray!20}
    Ours & 1 & 1 & \textbf{0.0322} & \textbf{0.0739}  \\
    \bottomrule[1pt]
    \end{tabular}}
    \vspace{-5mm}
\end{table}

\noindent\textbf{Number of CA and SA.}
We have analyzed the impact of varying the number of SA and CA on performance in Table \ref{tab:nyu_per}. We select the best setting for our GDNet.

\section{Conclusion}
In this paper, we study the limitations of current methods in compressed depth map super-resolution, identifying two primary challenges. 
First, bit-depth compression often simplifies depth representation in regions with subtle variations to a uniform level, making it challenging for existing models to accurately recover fine geometry details. 
Second, the presence of densely distributed noise in compressed depth maps can lead to inaccurate global geometric perception of the scene.
To address these issues, we propose a novel framework called GDNet for compressed depth map super-resolution. 
GDNet decouples the high-quality depth map reconstruction process by distinctly addressing global and detailed geometric feature learning. 
In addition, we introduce the fine geometry detail encoder (FGDE) to capture fine geometry details in high-resolution low-level image features while enriching them with complementary information from low-resolution context-level features. 
We also develop global geometry encoder (GGE) that constructs a compact feature representation in a low-rank space, effectively suppressing noise and extracting global geometric information.
The experiments demonstrate that GDNet achieves SoTA performance, surpassing existing methods in both fine geometry detail recovery and overall accuracy.

{
    \small
    \bibliographystyle{ieeenat_fullname}
    \bibliography{main}
}

\end{document}

%% file: preamble.tex
\usepackage[dvipsnames]{xcolor}
